\documentclass[a4paper, 10pt,twocolumn]{article}
\usepackage{latex12}
\usepackage{times}
\usepackage{balance}
\usepackage{amsmath,graphicx,float,tabularx}
\usepackage{booktabs}
\usepackage{multirow}
\usepackage{siunitx}

\usepackage[a4paper,
            hmargin=37pt,
            top=54pt,
            bottom=113pt,
            ]{geometry}

\begin{document}

\title{Efficient Hybrid Network: Inducting Scattering Features}

\author{Dmitry Minskiy, Miroslaw Bober\\
\emph{Centre for Vision, Speech and Signal Processing, University of Surrey}\\
\emph{\{d.minskiy, m.bober\}@surrey.ac.uk}  \\
}

\maketitle
\thispagestyle{empty}

\begin{abstract}
Recent work showed that hybrid networks, which combine predefined and learnt filters within a single architecture, are more amenable to theoretical analysis and less prone to overfitting in data-limited scenarios. However, their performance has yet to prove competitive against the conventional counterparts when sufficient amounts of training data are available. In an attempt to address this core limitation of current hybrid networks, we introduce an Efficient Hybrid Network (E-HybridNet). We show that it is the first scattering based approach that consistently outperforms its conventional counterparts on a diverse range of datasets. It is achieved with a novel inductive architecture that embeds scattering features into the network flow using Hybrid Fusion Blocks. We also demonstrate that the proposed design inherits the key property of prior hybrid networks - an effective generalisation in data-limited scenarios. Our approach successfully combines the best of the two worlds: flexibility and power of learnt features and stability and predictability of scattering representations. 
\end{abstract}

\Section{Introduction}

In this work, we define a hybrid network as a Convolutional Neural Network (CNN) that in its data flow combines learnable and predefined, i.e. non-learnable, filters. The vast majority of current hybrids applied to image classification problems employ a structure where scattering features replace a number of initial layers of a backbone CNN forming a "stack" of predefined non-learnable features followed by learnable filters. Although such a hybrid approach is superior in data-limited scenarios, in data-rich situations it tends to underperform conventional methods \cite{minskiy2021scatReview}.

An integral part of any hybrid design is a bank of predefined filters. Scattering transform \cite{mallat2012scatTransform} is the most commonly used for that method. Its design provides invariance to main sources of image variabilities such as translations and perspective deformations. The suitability of scattering features for image classification tasks has been proven both mathematically \cite{brunaMallat2013invariant_scatSVM} and empirically \cite{oyallonMallat2015deepRotoTransl, oyallon2017firstHybrid, oyallon2018firdtOrderScat}. However, the biggest challenge yet has been building data-adaptive invariants. Existing scattering-based approaches fail to produce representations that preserve the necessary information to outperform conventional CNN methods.

A number of hybrid networks were introduced based on various CNN backbones and different types of scattering networks \cite{singh2017dtcwScat, oyallon2018scatAnalsysis, cotter2019learnableScat}. Most architectures, share a similar stacking design paradigm, i.e. replace early convolutional layers with hand-crafted features. However, such an approach makes them prone to issues such as relatively low performance and lack of adaptability \cite{minskiy2021scatReview}. 

To overcome these issues, we develop a novel inductive architecture. Instead of replacing parts on a network with predefined filters, we suggest a procedure of embedding scattering coefficients into the main data pipeline to allow a network to select the most relevant features at any time during the training process. Feature embedding is achieved with Hybrid Fusion Blocks that combined with a backbone EfficientNet form the E-Hybrid network as detailed in section \ref{sec:building_an_e-hybrid_network}. We show that our design is the first scattering based hybrid approach to consistently outperform its conventional counterparts (section \ref{sec:hybrid_performance}). We demonstrate that our scattering feature embedding procedure is indeed the driving factor behind the success of the proposed architecture (section \ref{sec:Effect_of_Scattering_Features}), aids CNN generalisation and improves the overall performance in data-limited scenarios (section \ref{sec:generalsiation}). The key contributions of this work can be summarised as follows:

$\bullet$ We introduce a novel scattering-based hybrid architecture (E-HybridNet), which fuses information extracted by wavelet and scaling filters with conventional, learnt image features.

$\bullet$ We perform an extensive evaluation on diverse image classification datasets and demonstrate that the proposed hybrid architecture consistently outperforms the equivalent conventional networks. 

$\bullet$  Via the ablation studies, we gain a new understanding of the role of DropConnect, skip connections, batch normalisation in hybrid fusion, which helps us to identify the optimal design of the Hybrid Fusion Blocks.
  
$\bullet$ We demonstrate that Hybrid Fusion Blocks facilitate CNN generalisation with scattering representations and provide consistent performance gains, including for unbalanced datasets, with full or significantly reduced training sets. 

\begin{figure*}[!ht]
\centerline{\includegraphics[width=16cm]{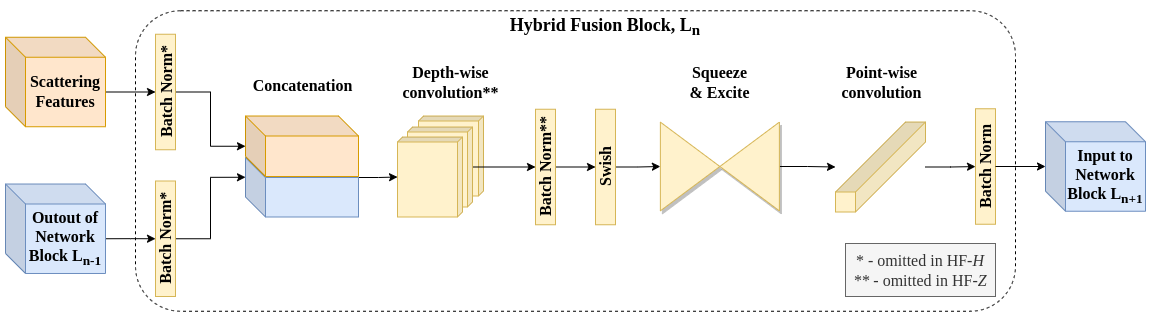}}
\caption{Hybrid Fusion Block architecture detailed.}
\label{fig:hybrid_blocks}
\end{figure*}
%
%

\section{Related Work}
\label{sec:related_work}
\textbf{Hybrid networks.} The idea originated in research presented in \cite{oyallon2017firstHybrid} where E. Oyallon et al. experimented with replacing initial layers of ResNet \cite{he2016resNet} and Wide ResNet \cite{zagoruyko2016wrn} with a scattering transform. This approach allowed deeper analysis and better interpretation of complex deep learning techniques and demonstrated that predefined filters grant theoretical guarantees required for building better networks and more stable representations. Presented results saw a sizeable improvement as compared to previous scattering-based approaches \cite{brunaMallat2013invariant_scatSVM}. Despite that, the performance with respect to fully-trained counterparts was still low. However, the promising properties of hybrid networks encouraged research to continue and alternative approaches were introduced. For example, Dual-Tree Complex Wavelet Transform-based network \cite{singh2017dtcwScat} or a Locally Invariant Convolutional Layer \cite{cotter2019learnableScat} which pioneered learning in the scattering domain. Although novel approaches including those reviewed in \cite{minskiy2021scatReview} saw further performance improvement, the issue of a significant performance gap between hybrid and fully-trained networks remained.

\textbf{Hand-crafted features.} A wide variety of predefined filters has been applied to hybrid designs, most of which are wavelet-based. Researchers employed both discrete and continuous wavelets including Haar \cite{fujieda2018waveletCNN}, Gabor \cite{czaja2019gaborHybrid}, and Morlet \cite{mallat2012scatTransform} wavelets. Later, it was demonstrated that wavelets are most efficient when they are used in a layered CNN-like configuration, such as Scattering Transform \cite{brunaMallat2013invariant_scatSVM} and Dual-Tree Complex Wavelet Transform \cite{selesnick2005dtcwt}.

Following the approach of E. Oyallon et al. \cite{oyallon2017firstHybrid}, we employ scattering transform, $S$, as the hand-crafted part for our hybrid network. It is formed by cascading wavelet transforms and modulus nonlinearity functions \cite{mallat2012scatTransform}. It guarantees translation invariance and linear response to deformations by separating the information along with multiple scales and orientations. As this work focuses on image classification, the wavelets are calculated from a zero-average complex wavelet $\psi$ as follows \cite{zarka2019scatDictonary}:

\begin{equation*}
    \psi_{j,\theta}(u) = 2^{-2j}\psi(2^{-j}r_{-\theta}u),
\end{equation*}
\begin{equation*}
    \psi_{j,\theta,\alpha} = Real(e^{-i\alpha}\psi_{j,\theta})
\end{equation*}

where $r_{-\theta}$ defines rotation, $2^j$ defines dilation, and $\alpha$ - phase shift.
In the proposed design we employ the first-order scattering coefficients which, for a given image $x$, are calculated by averaging rectified wavelet coefficients with a sub-sampling stride of $2^J$:
\begin{equation*}
Sx(u,k,\alpha) = |x\star\psi_{j,\theta,\alpha}|\star\phi_J(2^Ju); k=(j, \theta)
\end{equation*}
where non-linearity is obtained with a complex modulus function and $\phi_J$ is a Gaussian dilated by $2^J$ \cite{brunaMallat2013invariant_scatSVM} that eliminates the variations at scales smaller than $2^J$. Further details on the parameter selection are available in section \ref{sec:taining_details}.





\section{Building an Inductive Hybrid Network}
\label{sec:building_an_e-hybrid_network}

To address the weaknesses of existing hybrids, such as lacking performance and adaptability, we introduce an inductive approach as an alternative to the currently popular stacking paradigm. Conceptually, instead of replacing initial network stages, we enrich the backbone data flow along the entire network path with the output of a scattering transform. By design, this fusion allows scattering features to guide the training process, while the network can adaptively regulate their effect using the deep learning strategy. As a result, our architecture benefits from well-defined and informative hand-crafted features, as well as from an adaptable and flexible deep learning-based approach. The following two sections detail the building blocks of the proposed architecture and explain our design decisions.

\subsection{Hybrid Fusion Blocks}
\label{sec:hybrid_blocks}
Hybrid architecture furnishes a network with two complementary types of representations: fixed, hand-crafted with defined properties and learnt features that are adaptable to the data. The biggest challenge, then, is to fuse those while preserving the necessary information to maximise the classification performance. In our design, Hybrid Fusion Blocks enable that and steer the training procedure more effectively by drawing from the strengths of both types of features.

Hybrid Fusion Blocks are designed as encapsulated units and cause no additional disruption to the main network data flow. Figure \ref{fig:hybrid_blocks} illustrates how two streams of coefficients are embedded into a single tensor identical in shape to the output of the previous network layer. First, we perform feature expansion by concatenating scattering activations with an output of the previous layer, both of which are batch normalised. Then, inspired by the highly efficient MBConv block \cite{sandler2018mobilenetv2} we apply depth-wise convolution with a 3x3 kernel, followed by the Squeeze and Excitation (SE) stage. SE procedure allows the network to perform a re-adjustment through which it selectively emphasises informative features and suppresses less descriptive ones \cite{hu2018SEblock} enabling the network to prioritise scattering representations early in the process and regulate their effect as training progresses. Hybrid Fusion Block is concluded with point-wise convolution.

The above-described is the default Hybrid Fusion architecture (HF-$E$), we also introduce two alternatives to test the effects of batch normalisation during feature expansion and depth-wise convolution. This way, the HF-$H$ block has batch normalisation disabled in the first stage. While HF-$Z$ omits the depth-wise convolution procedure as shown in Figure \ref{fig:hybrid_blocks}.

Additionally, to assess the influence of  DropConnect \cite{wan2013dropConnect} and skip connection features when applied to the HF block, we evaluate three sub-variations for each of the three architectures. Option 0: without DropConnect and no skip connection; (1) no DropConnect with a skip connection; (2) with both DropConnect a skip connection enabled. Hence, the total number of block variations is nine, all of which are listed in Table \ref{tab:all_results}. 


\begin{table}[!b]
\newcolumntype{C}[1]{>{\hsize=#1\hsize\centering\arraybackslash}X}
    \begin{tabularx}{\linewidth}{ l c c }
      \hline
      Operator & Output Res & \# Channels \\
      \hline
       1. Conv3x3 & 112x112 & 32 \\
       2. MBConv1, 3x3 & 56x56 & 16 \\
       \textbf{3. HF-1, 3x3} & \textbf{56x56} & \textbf{24} \\
       4. MBConv6, 3x3 & 28x28 & 24 \\
       \textbf{5. HF-2, 3x3} & \textbf{28x28} & \textbf{40} \\
       6. MBConv6, 5x5 & 14x14 & 40 \\
       7. MBConv6, 3x3 & 14x14 & 80 \\
       8. MBConv6, 5x5 & 7x7 & 112 \\
       9. MBConv6, 5x5 & 7x7 & 192 \\
       10. MBConv6, 3x3 & 7x7 & 320 \\
       11. Conv1x1, Pooling, FC & 7x7 & 1280 \\
      \hline
    \end{tabularx}
    \caption{Overview of E-HybridNet-$E$ architecture for an input of resolution 224x224 pixels.}
    \label{tab:hybryd_dataflow}
\end{table}

\subsection{Scalable Network Architecture}
\label{sec:network_architecture}

Our feature embedding approach requires careful spatial feature resolution management, which restricts a selection of compatible backbone networks. The primary challenge is to align the two streams of features, network activations and scattering transform output, with respect to their spatial resolution.

One of the networks that meet this requirement is EfficnetNet \cite{tan2019efficientnet}. Scattering networks can be parameterised (as per section \ref{sec:taining_details}) to achieve alignment with respect to the resolution of the corresponding spatial features. Although other architectures previously employed with hybrids such as ResNet \cite{he2016resNet, oyallon2018firdtOrderScat}, Wide-ResNet \cite{zagoruyko2016wrn, oyallon2018scatAnalsysis} and VGG \cite{simonyan2014vgg, cotter2019learnableScat} also fulfill our resolution requirement, EfficientNet seem favorable due to its high performance-complexity trade-off, flexibility and wide recognition in applications related to image classification. 

 Previous research showed \cite{oyallon2018scatAnalsysis, cotter2019learnableScat} that scattering features are most effective when applied to early CNN stages. Hence, to build E-HybridNet we insert one Hybrid Fusion Block (HF-1) before original stage 3 of EfficientNet and another one (HF-2) just after it. Table \ref{tab:hybryd_dataflow} portrays the E-HybridNet-B0-$E$ architecture and demonstrates the integration of two Hybrid Fusion Blocks. We apply a two-step process to enhance the effect and form a flow of scattering features. 
 
 Importantly, due to the flexibility of the backbone network and adaptable HF block design, E-HybridNet is a highly scalable architecture. Unlike other existing hybrid networks, its complexity can easily be adjusted without re-configuring the predefined part.
 


\begin{table*}[!ht]
    \centering 
    \begin{tabular}{ l| c c c c }
      \hline
       Network Type & Caltech-256, \% & Flowers-102, \% & CoronaHack-2, \% & CoronaHack-3, \%\\
       \hline
        \textbf{EfficientNet-B0} & 54.23 & 94.63 & 92.58 & 85.66 \\
        E-HybridNet-B0-$E$0 & \textbf{60.24} & \textbf{97.75} & 95.10 & 87.73 \\
        E-HybridNet-B0-$E$1 & 60.08 & 96.52 & \textbf{95.49} & \textbf{87.82} \\
        E-HybridNet-B0-$E$3 & 58.92 & 97.63 & 95.21 & 86.27 \\
       \hline
        E-HybridNet-B0-$Z$0 & 59.50 & 97.12 & 94.09 & 84.24 \\
        E-HybridNet-B0-$Z$1 & \textbf{60.27} & 96.73 & 93.86 & \textbf{89.44} \\
        E-HybridNet-B0-$Z$3 & 57.63 & \textbf{97.63} & \textbf{94.32} & 85.82 \\
       \hline
        E-HybridNet-B0-$H$0 & \textbf{60.09} & 96.52 & \textbf{95.15} & 80.49 \\
        E-HybridNet-B0-$H$1 & 58.94 & 96.14 & 94.85 & \textbf{83.52} \\
        E-HybridNet-B0-$H$3 & 59.93 & \textbf{97.03} & 92.73 & 81.46 \\
       \hline
        \textbf{EfficientNet-B3} & 49.07 & 96.35 & 92.35 & 80.56 \\
        E-HybridNet-B3-$E$0 & 52.40 & 96.55 & \textbf{95.19} & 85.39 \\
        E-HybridNet-B3-$E$1 & 54.88 & 96.84 & 93.59 & 86.35 \\
        E-HybridNet-B3-$E$3 & \textbf{56.78} & \textbf{97.17} & 94.38 & \textbf{87.44} \\
       \hline
        E-HybridNet-B3-$Z$0 & \textbf{56.23} & \textbf{96.75} & \textbf{94.86} & 85.00 \\
        E-HybridNet-B3-$Z$1 & 52.53 & 96.96 & 93.22 & \textbf{86.68} \\
        E-HybridNet-B3-$Z$3 & 54.99 & 96.42 & 94.21 & 85.58 \\
       \hline
        E-HybridNet-B3-$H$0 & \textbf{55.83} & 96.95 & \textbf{93.31} & 83.93 \\
        E-HybridNet-B3-$H$1 & 54.75 & \textbf{97.53} & 92.63 & \textbf{85.07} \\
        E-HybridNet-B3-$H$3 & 53.75 & 96.91 & 92.38 & 84.87 \\
       \hline
    \end{tabular}
    \caption{Results for EfficientNet B0, B3 and hybrid networks based on those.}
    \label{tab:all_results}
\end{table*}


\section{Experimental Setup}
This section reviews key elements of the experimental setup and justifies the reasons behind certain decisions.

\subsection{Datasets}
\label{sec:datasets}
Datasets were selected based on the following principles: (i) data is representative of a real-world problem; (ii) to form a variety of tasks with generic and specific data; (iii) data is of relatively high image resolution
; (iv) the size of the dataset is not too large due to a considerable number of required experiments. Following is a brief introduction to each of the datasets employed. 

Flowers-102 \cite{nilsback2008flowers}: a topic-specific and structure intensive dataset. It consists of 6,652 training and 819 test images non-uniformly split across 102 classes. CoronaHack \cite{cohen2020coronaHack} is a medical dataset that consists of a mixture of CT scans and X-ray images. It features 5,309 training and 624 test images that can be split either into 2 (disease, no disease) or 3 (virus, bacteria and healthy) categories. We test our networks on both variations. Caltech-256 \cite{griffin2006caltech256} presents 256 imbalanced categories of real-life objects, it has a training set of 23,824 images and a testing set of 5,956.

\textbf{Evaluation metric.} As all datasets are imbalanced, the often-used accuracy metric would not be representative as it is highly sensitive to that. Hence, mean Average Precision (mAP) is employed as the main evaluation metric and all results presented in the following discussions will refer to mAP.

\subsection{Training Details}
\label{sec:taining_details}
\textbf{Training procedure.} The primary challenge was to create an environment to fairly compare a wide variety of networks. For this, we followed an approach suggested in \cite{minskiy2021scatReview} and employed a standard training procedure across all experiments. To remove the optimisation bias, we trained all networks from scratch and kept most parameters constant. These include cross-entropy loss function, cosine-annealing scheduler with a step size equal to the number of epochs and SGD optimisation strategy. A minority of parameters varied depending on the dataset, such as the number of epochs, batch size and initial learning rate. More details are available in the project's GitHub directory: https://github.com/dminskiy/EHybridNet-icpr2022.

\textbf{Scattering features.} To ensure the scattering coefficients align with the EfficientNet backbone in terms of the feature spatial size as per section \ref{sec:network_architecture} the following scattering configurations were employed. For the first Hybrid Fusion Block, HF-1, $J$ was set to 2, whereas for the second, HF-2, $J=3$. The common parameters for both layers are the scattering order of 1, 8 angles $\theta$ and 4 phases $\alpha$. Scattering coefficients are computed with a modified to allow multi-GPU support Kymatio \cite{andreux2020kymatio} package, please see the project's directory for details.


%
%
\begin{figure}[!b]
\centerline{\includegraphics[width=\linewidth]{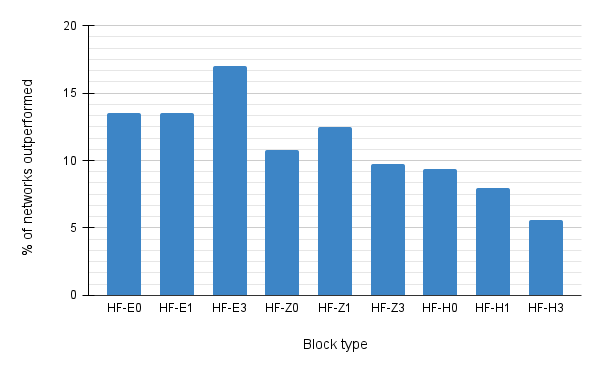}}
\caption{Performance comparison of Hybrid Fusion Block architectures.}
\label{fig:nets_outperformed_hybrids_B0&3}
\end{figure}
%
%


\begin{table*}[!ht]
    \centering 
    \begin{tabular}{ l| c c| c c| c c  }
      \toprule
      \multirow{2}{*}{Dataset} &
      \multicolumn{2}{c|}{All of training data} &
      \multicolumn{2}{c|}{50\% of training data} &
      \multicolumn{2}{c}{25\% of training data} \\
      & {Hybrid, \%} & {EfficientNet, \%} & {Hybrid, \%} & {EfficientNet, \%} & {Hybrid, \%} & {EfficientNet, \%} \\
      \midrule
       Caltech-256 & \textbf{60.24} & 54.23 & \textbf{54.60} & 46.68 & \textbf{45.81} & 38.03 \\
       Flowers-102 & \textbf{97.75} & 94.63 & \textbf{95.81} & 91.88 & \textbf{92.14} & 85.52 \\
       CoronaHack-2 & \textbf{95.10} & 92.58 & \textbf{94.93} & 90.85 & \textbf{92.70} & 88.92 \\
       CoronaHack-3 & \textbf{87.73} & 85.66 & \textbf{85.79} & 85.26 & \textbf{79.32} & 78.42 \\
      \bottomrule
    \end{tabular}
    \caption{Results with limited training data for E-HybridNet-B0-$E$0 and EfficientNet-B0.}
    \label{tab:limited_data_results}
\end{table*}


\section{Evaluation and Analysis}

In this section, we first explore the properties of the E-HybridNet, then evaluate its performance against the baseline EfficientNet. Finally, we assess the effect of scattering features on the network's efficiency in normal and data-limited conditions.  

\subsection{E-HybridNet Performance}
\label{sec:hybrid_performance}

\textbf{Hybrid Fusion Block architecture}.
First, we evaluate the effects of different design choices on the overall network performance, concentrating on the comparison of three primary design groups (HF-$E$, HF-$Z$ and HF-$H$) as per section \ref{sec:hybrid_blocks}. The impact of skip connections and DropConnect is considered later in the section.

To select the best architecture, we count the number of networks each of the candidate hybrid designs has outperformed and accumulate it across our four evaluation datasets and two network complexity configurations (B0 and B3), results are shown in figure \ref{fig:nets_outperformed_hybrids_B0&3}. To select the best performing design, we group the obtained results based on Hybrid Fusion Block variations ($E$, $Z$ or $H$). Thereby, we observe that HF-$E$ based networks outperformed 44.1\% of other hybrids, followed by HF-$Z$ design (better in 32.99\% cases) and HF-$H$ (22.92\%). These findings indicate the importance of batch normalisation before the features are concatenated, as this is the only difference between the leading HF-$E$ and the least effective HF-$H$ designs. Similarly, the depth-wise convolution stage proved effective as it allowed HF-$E$ architecture to be over 10\% more efficient than the HF-$Z$ design that does not have this feature.

\textbf{Hybrid Fusion Block variations.}
DropConnect and skip connection modules are present within the original EfficientNet architecture and play an important role in its success. Here, we analyse their effect on the E-HybrdNet, for this we apply the same method as before and count the number of peer-hybrids each individual architecture outperforms.

Figure \ref{fig:nets_outperformed_hybrids_B0&3} presents the aggregate result between B0 and B3-based networks and shows that the effect of these features depends on the fusion block architecture. The skip connection module, on average, did not change the performance of Hf-$E$ networks, for both HF-$E$0 and HF-$E$1 the number of outperformed networks remained the same, at 13.54\%. HF-$Z$ hybrids, on the other hand, benefited from the addition of the skip connection - the success rate improved by 1.74\%. Whereas, HF-$H$ architecture saw a drop in performance with the addition of this feature (-1.39\%). As for the DropConnect, the presence of this module took HF-$E$3 design to 17.01\%, the highest amongst all networks. However, its effect on other architectures was negative, reducing the performance by 2.78\% and 2.43\% for HF-$Z$ and HF-$H$ respectively.

In summary, we observe that DropConnect and skip connection modules can improve the adaptability of the network and they are most effective with the base E-hybridNet architecture - HF-$E$.

\textbf{Comparison with EfficientNet.}
The major drawback of previous hybrid architectures is their relatively low classification performance as compared to their conventional counterparts. Hence, in this experiment, we compare the performance of the base hybrid architecture (HF-$E$0) against the vanilla EfficientNet in B0 and B3 configurations applied to four diverse datasets.

Results presented in Table \ref{tab:all_results} evidence the dominance of the hybrid that is on average 3.43\% more accurate than the reference network with a maximum gap of over 6\% in B0 setup. When the backbone complexity is increased to B3, the average hybrid's advantage is 2.8\% with a maximum of 4.83\% To the best of our knowledge, it is the first time a hybrid network could consistently outperform the baseline network on a variety of tasks when entire training data is available.

\begin{figure}[!b]
\centerline{\includegraphics[width=\linewidth]{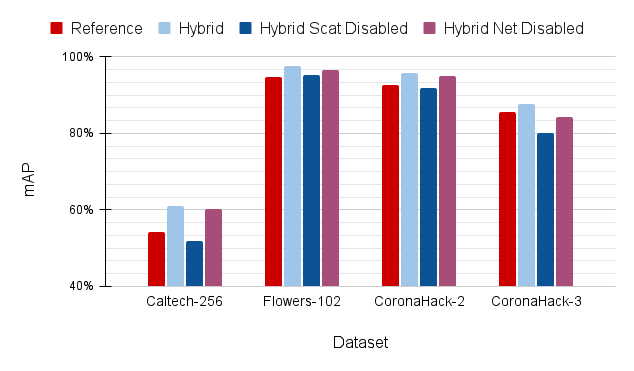}}
\caption{Results of the ablation study.}
\label{fig:deactivation}
\end{figure}

\subsection{Ablation Study}
\label{sec:Effect_of_Scattering_Features}
Here, we evaluate the contribution of scattering features to the performance of our hybrid design. As our HF block fuses scattering coefficients with an output of the corresponding network layer (network features, in short), an idea of the ablation study, therefore, is in disabling parts of the Hybrid Fusion Block. In one case, the scattering features are omitted from the merging stage while, in the other, the network features fed into the block will be ignored. As before, we use HF-$E$0 hybrid configuration for evaluation.

Figure \ref{fig:deactivation} summarises the results of this experiment across four datasets and following networks: EfficientNet-B0 (Reference), E-HybridNet-B0-$E$0 (Hybrid), E-HybridNet-B0-$E$0 with scattering features omitted (Hybrid Scat Disabled) and E-HybridNet-B0-$E$0 with network features ignored (Hybrid Net Disabled). Across the four datasets, the average loss in performance due to the lack of scattering features is 5.08\%, while the average performance drop due to the network features not being present is 1.48\% as compared to the baseline hybrid result. This indicates the significance of scattering features in hybrid architecture and shows that they are, indeed, the driving factor for its success. Additionally, we note that the average performance loss caused by disabling both hybrid and networks features (3.64\%) is approximately equal to the average gain of the hybrid versus the EfficientNet-B0 (3.76\%) architecture. This can be interpreted as "disabling" both parts of the Hybrid Fusion Block (or effectively removing the hybrid part of the network) is the same as running the standard EfficientNet-B0. Importantly, most of the loss when moving from hybrid to standard design is due to a lack of scattering features.

\subsection{Generalisation}
\label{sec:generalsiation}
Historically, hybrids' ability to generalise from small amounts of data have been their strongest advantage over their fully-trained counterparts. To validate that our design preserves this strength, we compare the base hybrid network (HF-$E$0) and its baseline EffcientNet-B0 in data-limited scenarios. We use our four evaluation datasets with 50\% and 75\% of the training samples removed randomly.

From results presented in Table \ref{tab:limited_data_results}, we observe that the hybrid network was constantly ahead of EfficientNet-B0. On average, E-HybridNet was 3.43\%, 4.11\% and 4.77\% more accurate than the reference architecture with full, half and quarter of the training set available. It corresponds to approximately 0.7\% of relative gain for the hybrid as the amount of data is halved. Hence, we conclude that our fusion procedure embeds scattering features effectively, aids CNN generalisation and improves overall performance in data-limited scenarios.

\section{Conclusions}

In this work, we introduced the E-Hybrid network, the first scattering based hybrid network that consistently outperformed its conventional counterparts. The evaluation showed that our design improves mAP by up to 6\% with an average gain of 3.11\% across tested datasets.
We also presented Hybrid Fusion Blocks, the integral part of the novel design. They allow a network to benefit from the adaptability of CNNs while being guided by powerful hand-crafted representations.
Our analysis showed that scattering features were indeed the driving force behind the success of the proposed hybrid design and constituted a major part of the performance gain. They also allowed E-HybridNets to be superior in data-limited scenarios, considerably improving the network's generalisation.


\textbf{Current limitations and further work.} Despite the advantages of the novel architecture, certain aspects of it provide further research opportunities. For instance, currently, each Hybrid Fusion Block requires recalculation of scattering features which slows inference and training speeds by around 60\%. Another important area that has not been fully addressed yet is training optimisation for E-Hybrid networks, so far they have been tested mostly under a common setup strategy to facilitate the concept verification and enable a fair comparison. There is also a number of scattering and backbone architectures that could be explored within the proposed architectural paradigm. Therefore, we believe that this work is only a promising start in exploring a wide range of research opportunities in hybrid networks.

\bibliographystyle{latex12}
\bibliography{latex12}

\end{document}